# Patch Based Classification of Remote Sensing Data: A Comparison of 2D-CNN, SVM and NN Classifiers

*Mahesh Pal[1*], Akshay[1], Himanshu Rohilla[2] and B. Charan Teja[1]*

[1]Department of Civil Engineering, NIT Kurukshetra,
[2]Department of Computer Science, NIT Kurukshetra, Haryana, India
[*]Corresponding author: E-mail: mpce_pal@yahoo.co.uk

**ABSTRACT**

Pixel based algorithms including back propagation neural networks (NN) and support vector machines (SVM) have been widely used for remotely sensed image classifications. Within last few years, deep learning based image classifier like convolution neural networks (2D-CNN) are becoming popular alternatives to these classifiers. In this paper, we compare performance of patch based SVM and NN with that of a deep learning algorithms comprising of 2D-CNN and fully connected layers. Similar to CNN which utilise image patches to derive features for further classification, we propose to use patches as an input in place of individual pixel with both SVM and NN classifiers. Two datasets, one multispectral and other hyperspectral data was used to compare the performance of different classifiers. Results with both datasets suggest the effectiveness of patch based SVM and NN classifiers in comparison to state of art 2D-CNN classifier.

## 1. INTRODUCTION

Land cover classification is one the most researched area using remote sensing data over last three decades. Out of various classification algorithms, back-propagation neural network (NN) has extensively been used by remote sensing community. With the introduction of support vector machines (SVM; [1]) and random forest classifiers [2], researchers shifted their focus to these classifiers due to their improved performance in terms of classification accuracy and ease of use. All of these classifiers were using the spectral information without giving any consideration to spatial information surrounding the individual pixels. Research in the area of spatial-spectral classification of remote sensing data [3] suggest the usefulness of spatial/contextual information in terms of reduced uncertainty in pixel labelling leading to a reduced salt and pepper noise in the classified image. With the introduction of deep learning (DL) algorithms in last ten years, remote sensing community started using these algorithms for land cover classification due to their capabilities in extracting features from image patches to achieve higher level of spatial information through feature extraction [4-8]. Out of several DL algorithms, CNN based deep learning architectures are found to be most powerful due to its computational efficiency and ability to recognize the 2-D patterns in the images [9]. The use of CNN has been extensively reported with remotely sensed images because of their ability to extract spatial-spectral features [10-15]. On the other hand, recent study by [16] suggest improved performance by patch based SVM in comparison to K-nearest Neighbors, linear discriminant analysis, naive Bayes and decision tree classifiers for hyperspectral datasets. Keeping in view of improved performance of patch based SVM, this paper compares the performance of 2-D CNN with that of patch based SVM and NN classifiers using one multispectral and other hyperspectral datasets. Proposed patch based SVM and NN classifiers consider both spectral as well as spatial/contextual information around a pixel. Basic assumption in considering neighbouring pixels during classification is that these pixels are having higher possibilities of belonging to same class.

## 2. DATASET

Dataset from two study areas located near the town of Littleport in eastern England (ETM+; Multispectral) and Anand district of Gujarat state, India (AVIRIS-NG; Hyperspectral) were used. For the Littleport area, ETM+ image was acquired on 19 June 2000 and a sub-image of size 307 columns by 330 rows covering the area of interest was used for classification. The classification problem involves the identification of seven land cover types (wheat, sugar beet, potato, onion, peas, lettuce and beans). AVIRIS-NG airborne campaign was carried out on 7-8 February 2016 over Anand district covering the research farms of Anand Agricultural University and the adjoining agricultural area. For AVIRIS-NG sensor, data was collected in 425 wavebands covering 350 to 2500 nm range at 4m spatial resolution. Major crops in this area during the campaign period were wheat at different stage, mustard, chickpea, Oat, Amaranth, potato, tomato, brinjal, chilli, different cultivars of tobacco at different stages, fodder maize at different stage, pearl millet, cabbage, fodder sorghum, amaranthus and fennel. A total of 13 classes were used to classify the AVIRIS-NG data. Out of total 425 bands, bands: 1-5,196-207, 285-320 was removed due to stripping problem in reflectance data. Thus a total of 372 bands covering a study area of 566 columns and 1101 rows were used for classification. For both datasets, ground references images were prepared after field visits.

## 3. CLASSIFICATION ALGORITHM

### 3.1. Convolution Neural Network

Convolutional neural network (CNN) comprises of a set simple processing unit arranged in layered structure with a weighted connection between each unit in a layer with every unit in adjacent layers.CNN based classifier consists of five layers: input layer, convolution layer, pooling layer, fully-connected layer (FC), and output layer [17].

The input layer of CNN uses image patches of predefined sizes extracted from the input image. Convolution layer transforms one set of feature maps into another set of feature maps using convolution with filters after passing through an activation function [18]. The input is fed in form of image patches of size m×n xB to the network, where *m* and *n* are the rows and columns of the patch and *B* denotes the number of bands. These input patches are then convolved using filters by CNN layer so as to produce feature maps or activation maps. Each convolution layer is followed by a pooling layer to reduce the spatial dimensions of the activation maps so as to reduce the computation task of successive layers. Various pooling layers are proposed in literature [9]. This process is repeated based on the number of convolution layers used in the study. As the choice of activation function is an important parameter in the design of DL algorithms, rectified linear units (ReLU) activation function [19] was used throughout in this work.

The output of final convolution layer acts as input to fully connected (FC; multi-layer neural network) layer after flattening. Flattening converts an image to a column so as to act an input to multi-layer neural network. The output of the FC layer is fed to an output layer providing class score using SoftMax activation function [17]. The whole network training is performed using back-propagation approach.

### 3.2. SVM and NN

The SVM uses a standard quadratic programming optimisation technique to determine location of class boundaries. It is designed for two-class problems and selects the linear decision boundaries providing greatest margin between two classes [1]. For linearly non-separable datasets, the SVM select a hyperplane that maximises the margin as well as minimising a quantity proportional to the number of misclassification errors at the same time. A slack variable is introduced to relax the restriction that all training data of a class lie on the same side of the optimal hyperplane and the trade-off between margin and misclassification error is controlled by a regularization parameter (C). In case of non-linear decision boundaries, SVM works by projecting the input data onto a high-dimensional feature space to convert it to a linear classification problem in the new feature space. To deal with the high computational cost in high-dimensional feature space use of a kernel function, satisfying the Mercer's theorem, was proposed [1].

The basic element of a NN is the processing node. Each processing node behaves like a biological neuron and performs two functions. First, it sums the values of its inputs. This sum is then passed through an activation function to generate an output. Any differentiable function can be used as an activation function, *f*. All the processing nodes in NN are arranged into layers, each fully interconnected to the following layer. There is no interconnection between the nodes of the same layer. In a NN, generally, there is an input layer that works as a distribution structure of the data being inputted to the network and not used for any type of processing. After this layer, one or more processing layers called the hidden layers follows. The final processing layer is called the output layer. A neural network with two or more hidden layers having large number of nodes and using sophisticated mathematical modeling [17], called as deep neural network is used in this study.

## 4. METHODOLOGY

To classify both datasets used in this study, spectral and spatial properties of pixels were used. For patch based SVM and NN, spatial properties of a pixel are extracted using neighbouring pixels from a window of fixed size (i.e. patch size). To train and test all three classifiers (2-D CNN, SVM and NN), images patches of different sizes (say p x p) were extracted using ground reference images for both multi- and hyperspctral dataset. While extracting a patch of size p, only those patches having central pixel as non-zero value (each class is assigned a number in reference images whereas area with no class information is assigned zero value) were considered during classification. For SVM, scikit-learn whereas Keras was used for 2D-CNN and NN implementation.

In case of SVM and NN, only pixel values of each patch from all wavebands were used as input to train and test these classifiers whereas the patches were processed by 2D-CNN to extract features and then passed to fully connected layers after pooling and flattening. Out of the total randomly selected samples, 75% data was used for training whereas remaining 25% for testing the classifiers. Various patch size values were considered with all classifiers to select a suitable value achieving best performance in terms of classification accuracy and classified image. The values of user-defined parameters for different classifiers obtained after large number of trials are provided in Table 1 for both datasets. For SVM, RBF kernel function was used and large number of trials was carried out to find optimal values of regularisation parameter (C) and $\gamma$ using a patch size of five. The performance of all classifier is measured in terms of overall classification accuracy.

## 5. RESULTS

Table 2 provides the results using both datasets in terms of classification accuracies by patch based SVM and NN classifiers as well as by 2D-CNN. With ETM+ dataset 2D-CNN provide higher accuracy than both SVM and NN classifiers but there is a marginal difference in classification accuracy between 2D-CNN and patch based SVM. For AVIRIS-NG dataset, SVM achieve highest accuracy out of all three classifiers and the difference in classification accuracy by all three classifier is very small. Variation in patch size is found to have profound effect on the accuracy of SVM classifier with both datasets whereas it has no major effect on the 2D-CNN and NN classifier. In case of ETM+ dataset, its accuracy declines to 61.13% from 90.74% when the patch size was increased from 1 to 11 whereas for AVIRIS-NG dataset it declines to 94.04% from 87.87% when patch size was increased from 1 to 13. In spite of increase in classification accuracy with increasing patch size with both 2D-CNN and NN classifiers, comparison of classified image indicates overlapping border regions, poorly defined boundary regions and classification of a smaller fields belonging to a class entirely into the other class with larger patch size. This suggests that one has to be careful in selecting a suitable patch size with both 2D-CNN and NN classifier.

Figures 1and 2 provides classified images of AVIRIS-NG and ETM+ datasets using different classifiers. A comparison of classified images also suggests improved performance by both patch based SVM and NN classifier with almost no salt and pepper noise but there is a clear different in the way classes are being assigned by all three classifiers with both ETM+ and AVIRIS-NG datasets.

Table 1. User defined parameters for different classifiers

| Classifier | ETM+ | AVIRIS-NG |
|---|---|---|
| 2-D CNN<br><br>Epoch= 2000, Batch size=128 | 2 convolution layers: 500, 100<br>2 Fully connected layers: 200, 84<br>Filter size=5x5,<br>Pooling size=2x2, Maxpooling Stride=2, Learning rate=0.001<br>Activation = Relu,<br>Optimiser= Adagrad | 2 convolution layers: 300, 200<br>2 Fully connected layers: 200, 84<br>Filter size=5x5<br>Pooling size=2x2, Maxpooling Stride=2, Learning rate=0.001<br>Activation = Relu<br>Optimiser= Adagrad |
| SVM | C=10, $\gamma$ =0.3 | C=30, $\gamma$ =3 |
| NN<br>Batch size=128 | Three Hidden layers: 500, 350, 150,<br>Learning rate=0.001,<br>Activation = Relu,<br>Epoch = 2000 | Three Hidden layers: 500, 350, 150<br>Learning rate = 0.001<br>Activation = Relu, Epoch=2000 |

Table 2. Classification accuracy by different classifiers

| Classifier | ETM+ | | AVIRIS-NG | |
|---|---|---|---|---|
| | Patch size | Accuracy (%) | Patch size | Accuracy (%) |
| 2D-CNN | 9 | 99.21 | 7 | 99.23 |
| SVM | 5 | 99.08 | 5 | 99.64 |
| NN | 5 | 98.30 | 5 | 99.59 |

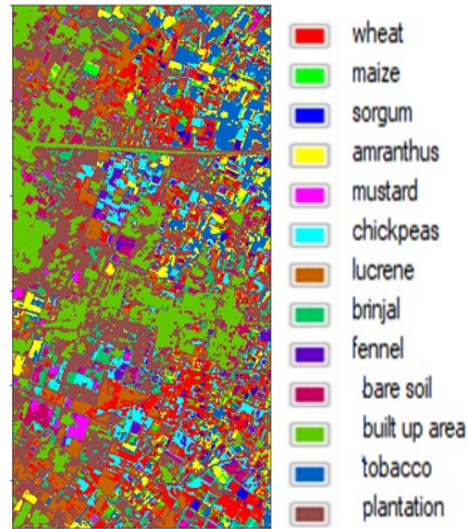

(a) 2D-CNN

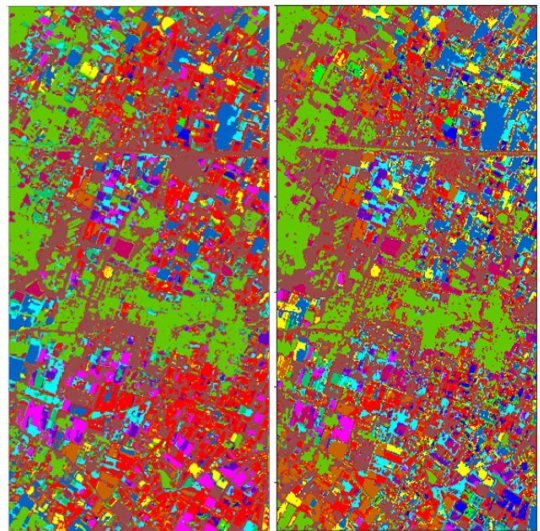

(b) SVM  (c) NN

Figure 1. Classified images of AVIRIS-NG dataset

## 6. CONCLUSIONS

This study proposes a classification approach using patch based information (i.e. spectral and spatial characteristics) using SVM and NN classifiers in comparison to 2D-CNN. A major conclusion is that both SVM and NN were able to improve classification accuracy significantly when used with spectral and spatial information obtained from a suitable patch size. The accuracy achieved is comparable to state of art 2D-CNN classifier with both datasets. Another conclusion is that size of patch has significant effect on the accuracy of SVM in comparison to 2D-CNN and NN classifiers. Finally, comparison of classified images suggest that different classifier is assigning different classes in a given area suggesting the need of further studies by collecting more ground data and using a part of it for classification whereas remaining for validation of classified images and accuracy.

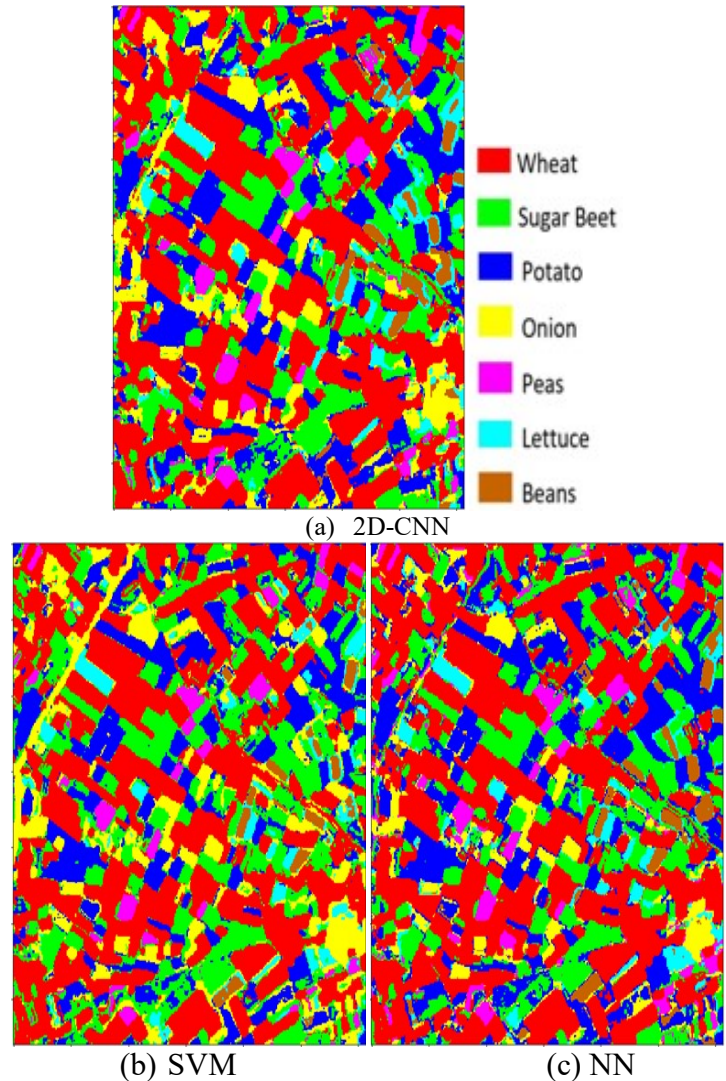

(a) 2D-CNN

(b) SVM  (c) NN

Figure 2. Classified images of ETM+ dataset